\documentclass[11pt, a4paper]{article}

\usepackage[utf8]{inputenc}
\usepackage[T1]{fontenc}
\usepackage{geometry}
\usepackage{graphicx}       
\usepackage{booktabs}       
\usepackage{hyperref}       
\usepackage{xcolor}         
\usepackage{listings}       
\usepackage{authblk}        
\usepackage{caption}
\usepackage{cite}           
\usepackage{tabularx}

\definecolor{codegreen}{rgb}{0,0.6,0}
\definecolor{codegray}{rgb}{0.5,0.5,0.5}
\definecolor{codepurple}{rgb}{0.58,0,0.82}
\definecolor{backcolour}{rgb}{0.95,0.95,0.92}

\lstdefinestyle{mystyle}{
    backgroundcolor=\color{backcolour},   
    commentstyle=\color{codegreen},
    keywordstyle=\color{magenta},
    numberstyle=\tiny\color{codegray},
    stringstyle=\color{codepurple},
    basicstyle=\ttfamily\footnotesize,
    breakatwhitespace=false,         
    breaklines=true,                 
    captionpos=b,                    
    keepspaces=true,                 
    numbers=left,                    
    numbersep=5pt,                  
    showspaces=false,                
    showstringspaces=false,
    showtabs=false,                  
    tabsize=2
}

\hypersetup{
    colorlinks=true,
    linkcolor=blue,
    filecolor=magenta,      
    urlcolor=cyan,
    citecolor=blue,
}

\title{FROAV: A Framework for RAG Observation and Agent Verification — Lowering the Barrier to LLM Agent Research}

\author[1]{Tzu-Hsuan Lin\thanks{Corresponding author: thlin40210@aethetech.com}}
\author[2]{Chih-Hsuan Kao\thanks{Corresponding author: carolyn@aethetech.com}}
\affil[1]{AetheTech\thanks{Website: www.aethetech.com}}
\date{\today}

\begin{document}

\maketitle

\begin{abstract}
The rapid advancement of Large Language Models (LLMs) and their integration into autonomous agent systems has created unprecedented opportunities for document analysis, decision support, and knowledge retrieval. However, the complexity of developing, evaluating, and iterating on LLM-based agent workflows presents significant barriers to researchers, particularly those without extensive software engineering expertise. We present FROAV (Framework for RAG Observation and Agent Verification), an open-source research platform that democratizes LLM agent research by providing a plug-and-play architecture combining visual workflow orchestration, a comprehensive evaluation framework, and extensible Python integration. FROAV implements a multi-stage Retrieval-Augmented Generation (RAG) pipeline coupled with a rigorous ``LLM-as-a-Judge'' evaluation system, all accessible through intuitive graphical interfaces. Our framework integrates n8n for no-code workflow design, PostgreSQL for granular data management, FastAPI for flexible backend logic, and Streamlit for human-in-the-loop interaction. Through this integrated ecosystem, researchers can rapidly prototype RAG strategies, conduct prompt engineering experiments, validate agent performance against human judgments, and collect structured feedback—all without writing infrastructure code. We demonstrate the framework's utility through its application to financial document analysis, while emphasizing its material-agnostic architecture that adapts to any domain requiring semantic analysis. FROAV represents a significant step toward making LLM agent research accessible to a broader scientific community, enabling researchers to focus on hypothesis testing and algorithmic innovation rather than system integration challenges.

\end{abstract}

\section{Introduction}

\subsection{The Promise and Challenge of LLM Agents}
Large Language Models have fundamentally transformed the landscape of natural language processing, demonstrating remarkable capabilities in text generation, reasoning, and knowledge synthesis \cite{brown2020, openai2023}. The emergence of LLM-based autonomous agents—systems that leverage LLMs to perceive, reason, and act upon complex information—has opened new frontiers in document analysis, decision support, and automated research assistance \cite{wang2024}. Particularly compelling is the application of Retrieval-Augmented Generation (RAG) techniques, which ground LLM outputs in external knowledge bases, significantly reducing hallucination and improving factual accuracy \cite{lewis2020}.

However, the development and evaluation of LLM agents remains a formidable challenge. Researchers face a complex technology stack requiring expertise in:

\begin{enumerate}
    \item \textbf{Workflow Orchestration}: Designing multi-step reasoning pipelines with appropriate error handling and state management.
    \item \textbf{Vector Database Management}: Implementing and optimizing semantic search over large document corpora.
    \item \textbf{Prompt Engineering}: Iteratively refining prompts across multiple agent roles and evaluation dimensions.
    \item \textbf{Evaluation Methodology}: Establishing rigorous metrics that capture both automated assessments and human judgments.
    \item \textbf{Data Infrastructure}: Managing persistent storage for experiments, logs, and feedback collection.
\end{enumerate}

This multifaceted complexity creates substantial barriers to entry, particularly for domain experts (e.g., financial analysts, medical researchers, legal scholars) who possess invaluable subject-matter expertise but lack software engineering backgrounds. The result is a concerning gap: those best positioned to validate LLM agent outputs in specialized domains often cannot participate meaningfully in system development and evaluation.

\subsection{The Need for Accessible Research Infrastructure}
Current approaches to LLM agent development typically require researchers to either:

\begin{enumerate}
    \item \textbf{Build from scratch}: Implementing custom pipelines using frameworks like LangChain or LlamaIndex, which demands significant programming expertise and system design knowledge.
    \item \textbf{Use closed platforms}: Relying on proprietary solutions that limit reproducibility, customization, and scientific transparency.
    \item \textbf{Cobble together tools}: Manually integrating disparate components (databases, APIs, frontends) with substantial integration overhead.
\end{enumerate}

None of these approaches adequately serve the research community's need for accessible, reproducible, and extensible experimentation platforms. The scientific study of LLM agents requires infrastructure that supports:

\begin{itemize}
    \item \textbf{Rapid prototyping} of different RAG strategies and prompt configurations.
    \item \textbf{Systematic evaluation} across multiple dimensions with both automated and human assessments.
    \item \textbf{Transparent logging} of all intermediate steps for debugging and analysis.
    \item \textbf{Flexible extension} to accommodate novel techniques and domain-specific requirements.
    \item \textbf{Collaborative workflows} enabling domain experts and ML researchers to contribute according to their strengths.
\end{itemize}

\subsection{Our Contribution: FROAV}
We introduce FROAV (Framework for RAG Observation and Agent Verification), an integrated research platform designed to lower the barrier to LLM agent experimentation while maintaining the flexibility required for serious scientific inquiry. FROAV's key contributions include:

\begin{enumerate}
    \item \textbf{Visual Workflow Orchestration}: Integration with n8n enables researchers to design, modify, and experiment with complex agent workflows through an intuitive drag-and-drop interface, eliminating the need for boilerplate infrastructure code.
    \item \textbf{Multi-Dimensional Evaluation Framework}: A comprehensive ``LLM-as-a-Judge'' system that evaluates agent outputs across four theoretically-grounded dimensions (Reliability, Completeness, Understandability, and Relevance), with multi-model consensus mechanisms to improve evaluation robustness.
    \item \textbf{Human-in-the-Loop Integration}: A Streamlit-based frontend that enables domain experts to review agent outputs, provide structured feedback, and contribute human judgments that can be correlated with automated assessments.
    \item \textbf{Granular Data Management}: PostgreSQL-based storage architecture that captures execution traces, evaluation results, and human feedback with full provenance tracking, enabling sophisticated post-hoc analysis.
    \item \textbf{Extensible Python Backend}: FastAPI-based services that expose clean interfaces for custom preprocessing, analysis routines, and integration with existing Python-based ML pipelines.
    \item \textbf{Containerized Deployment}: Docker Compose configuration enabling reproducible deployment across different computing environments with minimal setup.
\end{enumerate}

While we demonstrate FROAV through its application to financial document analysis (SEC 10-K, 10-Q filings), the architecture is deliberately material-agnostic, adaptable to any domain requiring systematic semantic analysis and agent verification. \footnote{\url{https://github.com/tw40210/FROAV_LLM}}

\begin{table}[htbp]
    \centering
    \caption{Development Effort Comparison: Manual Coding vs. FROAV Framework}
    \label{tab:barrier_comparison}
    \begin{tabularx}{\textwidth}{l X X}
        \toprule
        \textbf{Metric} & \textbf{Manual Coding} & \textbf{FROAV Framework} \\
        \midrule
        \textbf{Infrastructure Setup} & 40--50 hours (DB, Docker, API auth, Environment) & $\sim$1 hour (Single Docker Compose) \\
        \addlinespace
        \textbf{Workflow Logic} & 1000+ Lines of Python (Orchestration, Error handling) & 0 Lines (Visual Drag-and-Drop) \\
        \addlinespace
        \textbf{Evaluation Logic} & 1000+ Lines (Multi-model consensus, Persistence) & Pre-configured "Judge" nodes \\
        \addlinespace
        \textbf{HITL Interface} & 80+ hours (Frontend dev, API integration) & 2 hours (Modular Streamlit config) \\
        \addlinespace
        \textbf{Learning Curve} & High (Requires Senior Software Engineering) & Low (Domain Expert / Analyst friendly) \\
        \bottomrule
    \end{tabularx}
    \vspace{1mm}
    \small \textit{*Estimates based on typical development cycles for a production-grade RAG pipeline with evaluation.}
\end{table}

\section{Related Work}

\subsection{Retrieval-Augmented Generation}
RAG systems have emerged as a primary approach for grounding LLM outputs in external knowledge. The foundational work by Lewis et al. \cite{lewis2020} demonstrated that combining retrieval mechanisms with generative models significantly improves knowledge-intensive NLP tasks. Subsequent developments have refined RAG architectures through improved retrieval mechanisms \cite{izacard2021}, multi-hop reasoning \cite{trivedi2023}, and adaptive retrieval strategies \cite{jiang2023}.

However, implementing production-quality RAG systems remains challenging. Frameworks like LangChain \cite{harrison2022} and LlamaIndex \cite{liu2022} provide building blocks but require substantial programming expertise. Recent work has highlighted the sensitivity of RAG systems to retrieval quality, chunk sizing, and prompt design \cite{gao2023}, underscoring the need for systematic experimentation infrastructure.

\subsection{LLM-as-a-Judge Evaluation}
The ``LLM-as-a-Judge'' paradigm has gained traction as a scalable approach to evaluating LLM outputs. Zheng et al. \cite{zheng2023} demonstrated that strong LLMs can provide evaluations highly correlated with human judgments across various dimensions. This approach has been extended to domain-specific evaluation \cite{fu2023}, multi-dimensional assessment \cite{liuy2023}, and debate-based refinement \cite{du2023}.

Critical to effective LLM-as-a-Judge systems is addressing known biases, including position bias, verbosity bias, and self-enhancement \cite{zheng2023}. Multi-model consensus approaches have shown promise in mitigating these biases \cite{verga2024}, though implementation complexity has limited their adoption in research settings.

\subsection{Human-in-the-Loop Systems}
Human-in-the-loop (HITL) approaches remain essential for validating AI system outputs, particularly in high-stakes domains. Research has demonstrated the value of structured human feedback for improving model performance \cite{ouyang2022}, identifying failure modes \cite{ribeiro2020}, and establishing evaluation baselines \cite{clark2021}.

Tools like Label Studio and Argilla provide annotation interfaces, but these typically focus on training data collection rather than production system evaluation. Platforms designed for LLM evaluation often require significant integration effort and may not support custom evaluation dimensions.

\subsection{Research Infrastructure Gaps}
Despite these advances, significant gaps remain in the research infrastructure landscape:
\begin{enumerate}
    \item \textbf{Integration Burden}: Researchers must manually connect retrieval systems, LLMs, evaluation frameworks, and feedback collection tools.
    \item \textbf{Reproducibility Challenges}: Custom implementations often lack standardized logging, making experiments difficult to reproduce and compare.
    \item \textbf{Accessibility Barriers}: Most tools assume programming expertise, excluding domain experts from meaningful participation.
    \item \textbf{Evaluation Fragmentation}: Automated and human evaluations are typically collected through separate systems, complicating correlation analysis.
\end{enumerate}

\section{Methodology}
\begin{figure}[htbp]
    \centering
    \includegraphics[width=0.7\linewidth]{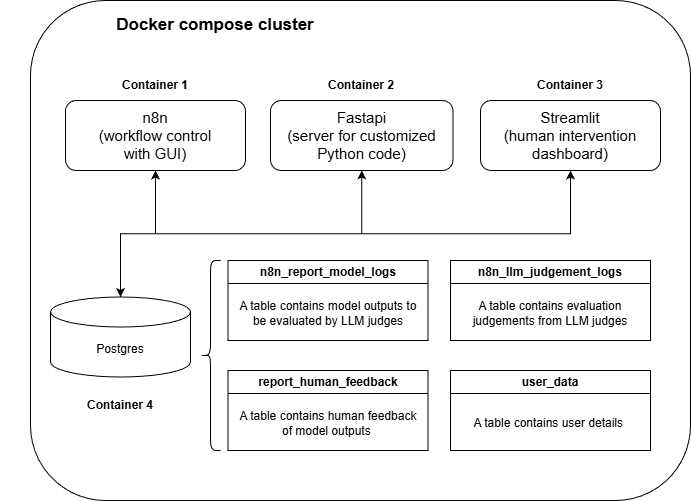}

    \caption{FROAV System Architecture showing the four primary components and their interconnections.}
    \label{fig:froav_arch}
\end{figure}

FROAV's architecture embodies a key design philosophy: \textbf{maximum accessibility with maximum flexibility}. We achieve this through a layered architecture that provides intuitive interfaces for common operations while exposing powerful extension points for advanced customization. Figure \ref{fig:froav_arch} illustrates the high-level system architecture.

\subsection{System Architecture Overview}
FROAV comprises four primary components orchestrated through Docker Compose for reproducible deployment:

\subsubsection{n8n Workflow Orchestrator}
n8n serves as the central nervous system of FROAV, providing visual workflow design capabilities that enable researchers to construct complex agent workflows without writing infrastructure code. Key capabilities include visual pipeline design, subworkflow composition, code injection points (JavaScript/Python), built-in AI nodes, and execution logging.

\subsubsection{PostgreSQL Data Layer}
PostgreSQL provides the persistent storage layer with a schema designed for research workflows for correlation analysis, temporal analysis, and access control.

\subsubsection{FastAPI Backend Services}
The FastAPI backend provides Python-based services that extend n8n's capabilities. This architecture enables researchers to implement custom preprocessing logic, integrate existing Python libraries, and extend the system with new endpoints without modifying core workflows.

\subsubsection{Streamlit Frontend}
The Streamlit frontend provides an accessible interface for human-in-the-loop evaluation, including an interactive report viewer, judgment browser, feedback collection, and user authentication.

\subsection{RAG Pipeline Architecture}
FROAV implements a sophisticated multi-stage RAG pipeline designed for complex document analysis tasks. The pipeline architecture supports iterative refinement based on evaluation feedback. While we use Supabase (PostgreSQL with pgvector) as the default third-party vector store service provider, FROAV is highly customizable.

\subsubsection{Core RAG Stages}
The RAG pipeline operates in three core stages:
\begin{enumerate}
    \item \textbf{PDF Content Parsing}: Utilizing custom FastAPI backend services to ingest PDF files and convert raw document content into structured text.
    \item \textbf{Materials Chunking}: Segmenting extracted text into contextually coherent chunks suitable for embedding.
    \item \textbf{Embedding and Vector Storage}: Transforming chunks into high-dimensional vectors and storing them with metadata in the vector database.
\end{enumerate}

\subsection{LLM-as-a-Judge Evaluation Framework}
FROAV implements a multi-model, multi-dimensional evaluation framework that addresses known limitations of single-model assessment.

\subsubsection{Evaluation Dimensions}
FROAV evaluates outputs across four dimensions (Table \ref{tab:dimensions}).

\begin{table}[htbp]
    \centering
    \caption{Evaluation Dimensions}
    \label{tab:dimensions}
    \begin{tabular}{lp{5cm}p{6cm}}
        \toprule
        \textbf{Dimension} & \textbf{Definition} & \textbf{Evaluation Focus} \\
        \midrule
        \textbf{Reliability} & Accuracy of stated facts & Data verification, calculation correctness, source fidelity \\
        \textbf{Completeness} & Presence of required information & Missing disclosures, omitted risks, incomplete analysis \\
        \textbf{Understandability} & Clarity of presentation & Jargon usage, logical structure, transparency \\
        \textbf{Relevance} & Decision-usefulness & Predictive value, confirmatory value, materiality \\
        \bottomrule
    \end{tabular}
\end{table}

Each dimension is evaluated by a specialized judge agent. For example, the Reliability judge utilizes the following system prompt:
\begin{quote}
    ``You are an expert financial analyst and auditor. Your primary task is to critically evaluate the Reliability of a given financial report. Approach the report with professional skepticism. Do not accept any data, calculation, or description at face value. Your objective is to verify, not just read.''
\end{quote}

\subsubsection{Multi-Model Consensus}
To mitigate individual model biases, each dimension is evaluated by multiple LLMs. The scores and corresponding rationales from various models are aggregated, and the median value is employed as the aggregated dimensional score to mitigate the influence of outlier models.

\subsection{Human-in-the-Loop Feedback System}
FROAV's feedback system enables systematic collection of human judgments aligned with automated evaluation dimensions. The database schema enables direct correlation between human and automated assessments, allowing researchers to identify biases, calibrate prompts, and study inter-annotator agreement.

\subsection{Extensibility and Customization}
FROAV's layered architecture supports multiple levels of customization. Researchers can modify workflows through the visual interface (swapping LLM providers, adjusting retrieval parameters) or extend functionality via the FastAPI server.

\section{Discussion}

\subsection{Accessibility Benefits}
FROAV significantly lowers barriers to LLM agent research for:
\begin{itemize}
    \item \textbf{Domain Experts}: Visual workflow design eliminates coding requirements.
    \item \textbf{ML Researchers}: Reduced boilerplate enables focus on algorithmic innovation.
    \item \textbf{Research Teams}: Shared infrastructure facilitates collaboration between experts and engineers.
\end{itemize}

\subsection{Flexibility Trade-offs}
FROAV's design explicitly balances accessibility and power (Table \ref{tab:tradeoffs}).

\begin{table}[htbp]
    \centering
    \caption{Flexibility Trade-offs}
    \label{tab:tradeoffs}
    \begin{tabular}{lp{3cm}p{3cm}p{3cm}}
        \toprule
        \textbf{Level} & \textbf{Accessibility} & \textbf{Capability} & \textbf{Use Case} \\
        \midrule
        Visual (n8n) & Very High & Standard workflows & Prompt iteration, model comparison \\
        Low-Code (JS) & High & Custom logic & Specialized aggregation \\
        Full-Code (FastAPI) & Medium & Unlimited & Novel preprocessing, custom models \\
        \bottomrule
    \end{tabular}
\end{table}

\subsection{Limitations and Future Work}
Current limitations include the restricted scope of the four-dimension framework and the lack of enterprise-scale stress testing. Future work involves developing an LLM-as-Judges methodology applied to financial SEC filings, calibrating LLM scoring with human experts using FROAV. This implementation is projected to significantly reduce the time required to achieve robust results.

\section{Conclusion}
FROAV represents a significant advancement in research infrastructure for LLM agent development and evaluation. By integrating visual workflow design, comprehensive evaluation frameworks, and human-in-the-loop feedback within a unified platform, FROAV democratizes access to sophisticated agent research capabilities.

\bibliographystyle{plain}

\begin{thebibliography}{99}

\bibitem{brown2020} Brown, T. B., et al. (2020). Language models are few-shot learners. \textit{Advances in Neural Information Processing Systems}, 33, 1877-1901.

\bibitem{clark2021} Clark, P., et al. (2021). TruthfulQA: Measuring how models mimic human falsehoods. \textit{arXiv preprint arXiv:2109.07958}.

\bibitem{du2023} Du, Y., et al. (2023). Improving factuality and reasoning in language models through multiagent debate. \textit{arXiv preprint arXiv:2305.14325}.

\bibitem{fu2023} Fu, J., et al. (2023). GPTScore: Evaluate as you desire. \textit{arXiv preprint arXiv:2302.04166}.

\bibitem{gao2023} Gao, Y., et al. (2023). Retrieval-augmented generation for large language models: A survey. \textit{arXiv preprint arXiv:2312.10997}.

\bibitem{harrison2022} Harrison, C. (2022). LangChain: Building applications with LLMs through composability. \textit{GitHub Repository}.

\bibitem{izacard2021} Izacard, G., \& Grave, E. (2021). Leveraging passage retrieval with generative models for open domain question answering. \textit{EACL 2021}.

\bibitem{jiang2023} Jiang, Z., et al. (2023). Active retrieval augmented generation. \textit{arXiv preprint arXiv:2305.06983}.

\bibitem{lewis2020} Lewis, P., et al. (2020). Retrieval-augmented generation for knowledge-intensive NLP tasks. \textit{Advances in Neural Information Processing Systems}, 33, 9459-9474.

\bibitem{liu2022} Liu, J. (2022). LlamaIndex: A data framework for LLM applications. \textit{GitHub Repository}.

\bibitem{liuy2023} Liu, Y., et al. (2023). G-Eval: NLG evaluation using GPT-4 with better human alignment. \textit{arXiv preprint arXiv:2303.16634}.

\bibitem{openai2023} OpenAI. (2023). GPT-4 Technical Report. \textit{arXiv preprint arXiv:2303.08774}.

\bibitem{ouyang2022} Ouyang, L., et al. (2022). Training language models to follow instructions with human feedback. \textit{Advances in Neural Information Processing Systems}, 35, 27730-27744.

\bibitem{ribeiro2020} Ribeiro, M. T., et al. (2020). Beyond accuracy: Behavioral testing of NLP models with CheckList. \textit{ACL 2020}.

\bibitem{trivedi2023} Trivedi, H., et al. (2023). Interleaving retrieval with chain-of-thought reasoning for knowledge-intensive multi-step questions. \textit{ACL 2023}.

\bibitem{verga2024} Verga, P., et al. (2024). Replacing judges with juries: Evaluating LLM generations with a panel of diverse models. \textit{arXiv preprint arXiv:2404.18796}.

\bibitem{wang2024} Wang, L., et al. (2024). A survey on large language model based autonomous agents. \textit{Frontiers of Computer Science}, 18(6), 186345.

\bibitem{zheng2023} Zheng, L., et al. (2023). Judging LLM-as-a-judge with MT-Bench and Chatbot Arena. \textit{arXiv preprint arXiv:2306.05685}.

\end{thebibliography}

\end{document}